\documentclass[conference]{IEEEtran}
\IEEEoverridecommandlockouts
\usepackage{cite}
\usepackage{amsmath,amssymb,amsfonts}
\usepackage{algorithmic}
\usepackage{graphicx}
\usepackage{textcomp}
\usepackage{xcolor}
\usepackage{subcaption}
\usepackage{booktabs}
\def\BibTeX{{\rm B\kern-.05em{\sc i\kern-.025em b}\kern-.08em
    T\kern-.1667em\lower.7ex\hbox{E}\kern-.125emX}}
\renewcommand{\footnoterule}{%
  \kern -3pt
  \hrule width 0.3\columnwidth height 0.4pt
  \kern 2.6pt
}

\begin{document}
\makeatletter
\newcommand{\linebreakand}{%
  \end{@IEEEauthorhalign}
  \hfill\mbox{}\par
  \mbox{}\hfill\begin{@IEEEauthorhalign}
}
\makeatother

\title{Optimize the Cob, Then the Kernel: Optimizing Beam Search for Generative Recommendation Inference}



\author{
\IEEEauthorblockN{
Shing Wai Pun\textsuperscript{\dag},
Kai Wu\textsuperscript{\dag},
Pengfei Xia,
Hui Zang,
Wei Guo,
Yong Liu
}
\IEEEauthorblockA{
Huawei Technologies Co., Ltd.
}
\thanks{\textsuperscript{\dag}Corresponding authors:
Shing Wai Pun (shingwaipun@gmail.com)
and Kai Wu (eewukai@gmail.com).}
\thanks{Code will be made publicly available.}
}

\maketitle

\begin{abstract}
Generative recommendation (GR) inference differs substantially from conventional LLM serving, featuring long input sequences, wide beam search, and only a few fixed output tokens. Existing LLM serving engines such as vLLM process beams as batched independent sequences, causing redundant KV cache accesses over the shared prompt prefix. Furthermore, vLLM executes each beam search generation step as a separate engine invocation, incurring host-side beam processing overhead between decoding steps and substantial runtime KV block management in PagedAttention. Using Tree Attention, we concatenate beams along the sequence dimension to eliminate redundant KV accesses, and further optimize execution through pre-generated partial-attention masks, static KV cache allocation, loop-unrolled generation dispatch, TorchAir graph compilation, automatic operator fusion, and vocabulary-constrained decoding. Without the use of custom kernels, our approach reduces end-to-end latency by up to 25$\times$ over vLLM on Ascend NPUs, and achieves generation latency comparable to a SOTA solution using custom kernels, showing that the generation pipeline should be optimized holistically before resorting to custom kernel development.
\end{abstract}


\section{Introduction}
Generative recommendation (GR) has recently emerged as an alternative paradigm to conventional retrieval- and ranking-based recommendation, formulating recommendation as a conditional sequence generation problem. As shown in Fig.~\ref{fig:gr_pipeline}, given a tokenized representation of a user's interaction history and contextual information, a generative recommendation model directly predicts recommended items as short sequences of discrete item tokens, commonly referred to as Semantic IDs (SIDs)~\cite{sidrajput2023}. Item recommendations are then generated autoregressively through beam search.

\begin{figure}[htbp]
    \centering
    \includegraphics[width=1.0\linewidth]{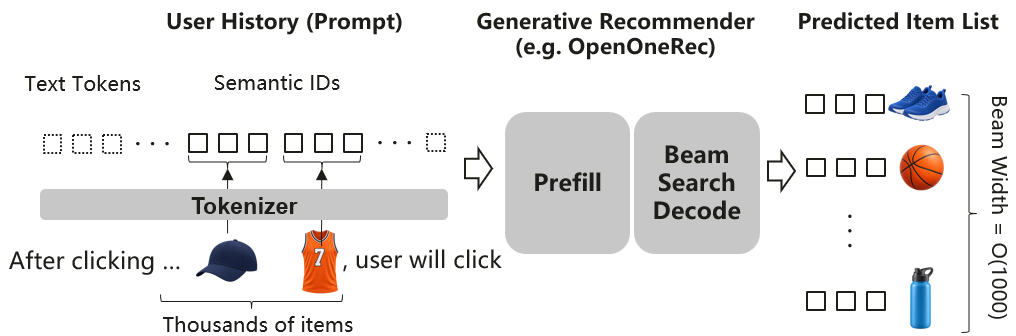}
    \caption{The generation pipeline of generative recommendation models.}
    \label{fig:gr_pipeline}
\end{figure}

Unlike conventional LLM inference, which typically uses greedy decoding or narrow-beam search to produce long, variable-length outputs, GR inference is characterized by extremely wide beam search and short, fixed-length outputs. To generate a diverse set of recommended items, the beam width can often exceed thousands, while each candidate typically consists of only a few output tokens (e.g., three SID tokens). Recommendations are also conditioned on long user interaction histories, resulting in long input sequences that are shared across all beams. Moreover, recommendation serving is highly latency-sensitive, typically operating under inference end-to-end latency budgets on the order of 10--100 ms.

Existing LLM serving engines, such as vLLM, are not designed for this unique combination of long shared prefixes, extremely wide beam search, and short, fixed decoding lengths. In particular, vLLM processes beams as batched independent sequences, causing redundant KV cache accesses over the shared prefix. Furthermore, each beam search generation step is executed as a separate engine invocation, incurring host-side processing overhead between decoding steps, and substantial runtime KV block management in PagedAttention over each generation step.

In this work, we design a GR inference pipeline that is explicitly optimized for GR's workload characteristics. Building upon Tree Attention, we concatenate beams along the sequence dimension and process them with a compile-time pre-generated tree attention mask that restricts token visibility between beams, eliminating redundant KV cache accesses over shared prefixes. We further exploit the short and fixed decoding length of GR by allocating a static-sized KV cache based on the known beam width and output length, thereby minimizing runtime memory-management overhead and removing inter-step host-side control. We also use TorchAir graph compilation, automatic operator fusion, and SID vocabulary-constrained decoding to further improve end-to-end latency.

Our evaluation shows that these system-level optimizations substantially improve GR inference latencies without the need for custom kernel development. Our approach reduces end-to-end latency by up to 25$\times$ compared with vLLM and achieves generation latency comparable to a SOTA GR serving solution that relies on custom operators. More broadly, our results serve as a reminder of Amdahl's law, where kernel-level speedups are limited by the costs of their surrounding operators. This motivates optimizing the full generation pipeline holistically, followed by targeted development of the individual kernels that remain bottlenecks---optimizing the ``cob'' first, then the kernel.

Our main contributions are:
\begin{itemize}
\item Building on Tree Attention, we develop a complete end-to-end inference pipeline and detail the system design choices used, including pre-generated partial-attention masks, static KV cache allocation, and token generation loop unrolling.

\item We implement and optimize the pipeline on Ascend NPUs, detailing how graph compilation, operator fusion, and execution path optimization are applied within the Ascend software stack.

\item We demonstrate up to 25$\times$ lower end-to-end latency than vLLM while achieving generation latency comparable to a SOTA implementation based on custom operators.
\end{itemize}

\section{Background and Motivation}

\subsection{Semantic IDs in OpenOneRec}
In OpenOneRec, each item is represented as a fixed-length sequence of three Semantic-ID (SID) tokens~\cite{tiger, plum, openonerec, unisgr}. These SIDs are obtained by applying three-layer residual-quantization K-means (RQ-Kmeans) to multimodal item embeddings. At each layer, K-means constructs a codebook, and the nearest centroid’s index becomes the corresponding token. Subtracting this centroid produces the residual passed to the next layer. The tokens therefore describe the item from coarse semantic categories (e.g., sports equipment) to finer distinctions (e.g., basketballs), allowing semantically similar items to share prefixes. The resulting three-token SID sequence serves as the item representation used for autoregressive recommendation.

\subsection{Beam Search in vLLM}
vLLM is an open-source inference and serving framework for LLMs, designed to support efficient generation through request batching and paged KV cache management~\cite{pagedattention}. In current versions of vLLM, beam search is implemented as a step-by-step loop outside the generation engine, repeatedly invoking \verb|generate()| with \verb|max_tokens=1|. This introduces three major inefficiencies for GR workloads.

\paragraph*{Bottleneck 1: Redundant KV cache accesses}

Within each decoding step, vLLM processes active beams as independent sequences, logically repeating the shared prefix across the batch dimension, as illustrated in Fig.~\ref{fig:treeatten} (lower-left). Attention is therefore evaluated separately for each beam's query against the same prefix KV states. This batched representation does not allow for the common prefix to be reused effectively across beam queries in the attention kernel, leading to redundant KV accesses that are exacerbated by the extremely wide beam search in GR.

Enabling automatic prefix caching in vLLM can reduce physical KV storage duplication by allowing beams to reuse cached prefix blocks. However, this doesn't change vLLM's attention kernel tiling strategy, which is unaware and therefore unable to take advantage of the shared prefix KV states across beam queries.

\paragraph*{Bottleneck 2: Runtime KV block management}
vLLM manages the KV cache using PagedAttention to support general LLM workloads with variable sequence lengths. Using a pre-determined KV block size (fixed to 128 tokens in vLLM Ascend), vLLM manages KV blocks during runtime through block allocation, prefix-cache block matching, block-table updates, block reference-count management, and block releases across each decoding step in beam search. This mechanism is suboptimal in GR workloads, leading to memory fragmentation, repeated block re-allocation and uncached KV suffix recomputation. As each beam is treated as a separate generation request, each beam is allocated a separate KV block for its beam-specific tokens. Since generation is limited to a few tokens, most of the capacity in these 128-token blocks remains unused, where the wasted capacity is multiplied by the large beam widths. Furthermore, automatic prefix caching only retains fully filled KV blocks across generation steps, so each partially filled beam-specific block is released and subsequently reallocated at the next generation step, requiring its previously computed KV states to be recomputed.

\paragraph*{Bottleneck 3: Host-side beam processing}

vLLM performs beam selection and processing on the host, requiring a large number of candidate token IDs and log probabilities to be transferred from the NPU after each decoding step. The host then updates beam scores and selects the surviving beams before generation can resume, introducing synchronization overhead and gaps where the NPU is idle.

\subsection{Existing GR Inference Systems}
xGR~\cite{xgr} is a serving system optimized for GR inference. Its custom xAttention operator separates the KV cache into a shared prompt region and beam-specific regions, allowing all beams to reuse the prompt KV states without access duplication. Attention over the two regions is computed separately and combined through online softmax to preserve the result of joint attention. After beam selection, dedicated cache-management operators reorder and replicate the beam-specific KV states according to the selected parent indices for the next decoding step.

UniSGR~\cite{unisgr} proposes Semantic Tree Attention with Reorganized KV Cache (STARK), which uses Tree Attention, like ours. Its key idea is to move beam expansion to the sequence dimension, using a precomputed tree attention mask to exploit shared prefixes while preserving beam search semantics. However, the paper provides limited details on the practical implementation and system-level optimizations.

\begin{figure*}[htbp]
    \centering
    \includegraphics[width=1.0\linewidth]{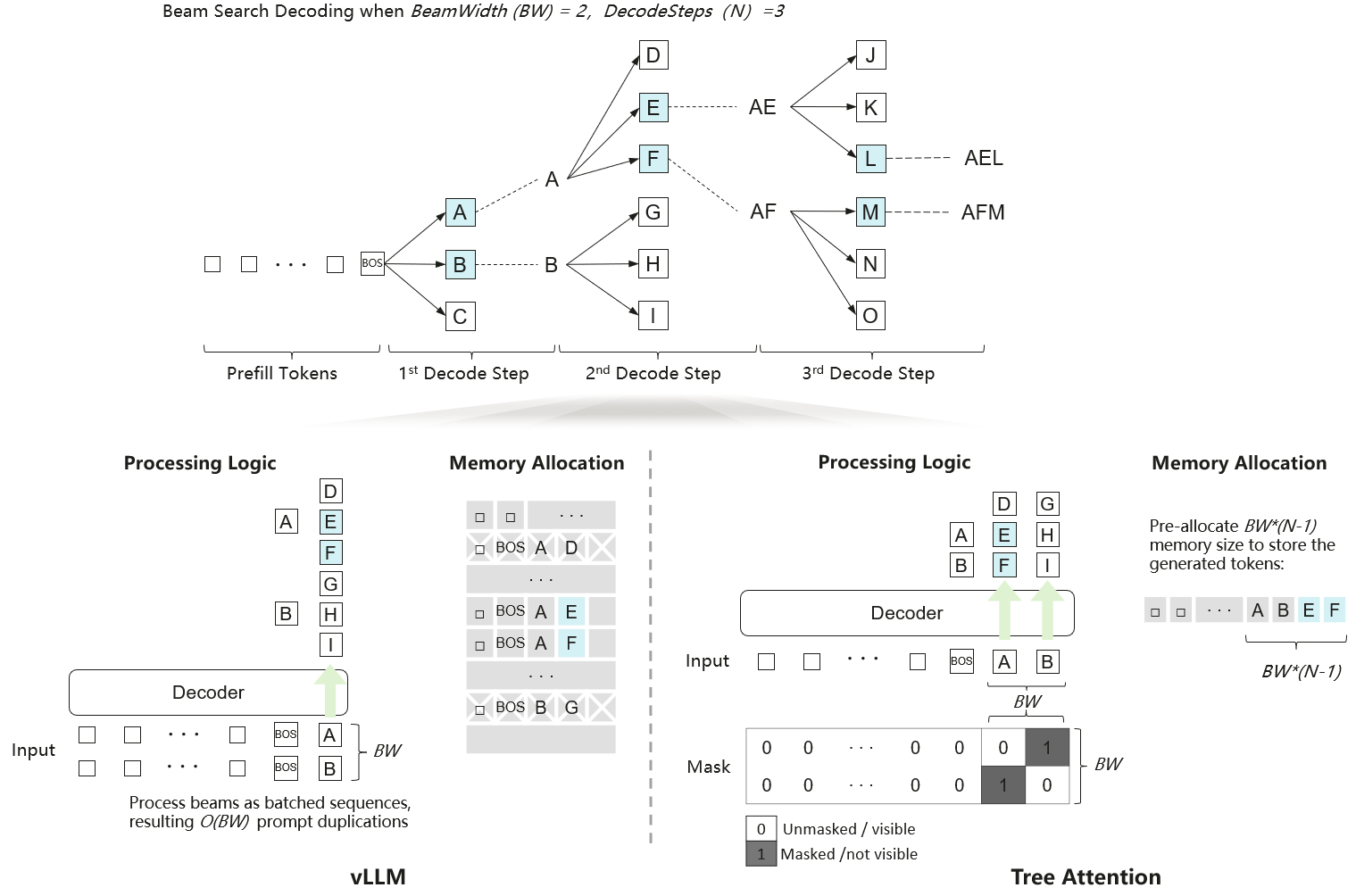}
    \caption{(Upper) Illustration of beam search with a beam width of two and three decode steps. (Lower-left) Processing logic and memory management in vLLM at the second decoding step. (Lower-right) Processing logic and memory management in Tree Attention at the second decoding step.}
    \label{fig:treeatten}
\end{figure*}


\section{Proposed Method}
\subsection{Tree Attention}
We address the first bottleneck of redundant KV cache accesses by applying Tree Attention to avoid replication of the shared prefix while preserving the logical structure of beam search. As opposed to placing beams along the batch dimension in vLLM, the Tree Attention implementation appends the tokens selected at each decoding step contiguously \textbf{along the sequence dimension} to a single instance of the prompt. This evaluates all the beams together with the prompt as one long sequence. Figure~\ref{fig:treeatten} illustrates this process for a beam width of 2 in the second decoding step. Tokens E and F are selected as the top-$BW$ tokens for the second decode step, and are consequently flattened and appended to the input sequence for the next decoding step.

The use of a flattened tree structure requires a corresponding attention mask that only allows each generated token to attend to itself, its ancestors, and the shared prompt, but not to tokens in other beam branches.
The attention operation can be expressed mathematically as:
\begin{equation}
\operatorname{softmax}
\left(
\frac{\mathbf{Q}\mathbf{K}^{\mathsf{T}}}{\sqrt{d}}
-\lambda
\mathbf{M}
\right)
\mathbf{V}, ~~\lambda \rightarrow +\infty. \footnote{In implementation, the binary mask \textbf{M} is directly passed to the fused FlashAttention operator.}
\end{equation} 
where $\mathbf{Q}$, $\mathbf{K}$ and $\mathbf{V}$ denote the query, key and value matrices. The attention mask $\mathbf{M}$ controls the visibility between tokens, and its elements are defined as
\begin{equation}
M_{i,j}
=
\begin{cases}
0~(\text{visible}), & j \in \mathcal{P}\cup\operatorname{Path}(i), \\
1~(\text{masked}), & \text{otherwise},
\end{cases}
\end{equation}
where $\mathcal{P}$ denotes the positions of the shared prompt, and $\operatorname{Path}(i)$ denotes the ancestor path of token $i$, including  itself. With this mask mechanism, each token observes the same attention causality as it would in an independently evaluated beam, while all beams are evaluated in a single attention operation.

\subsection{Partial Mask Pre-Generation}

\begin{figure*}[htbp]
\centering
\includegraphics[width=\linewidth]{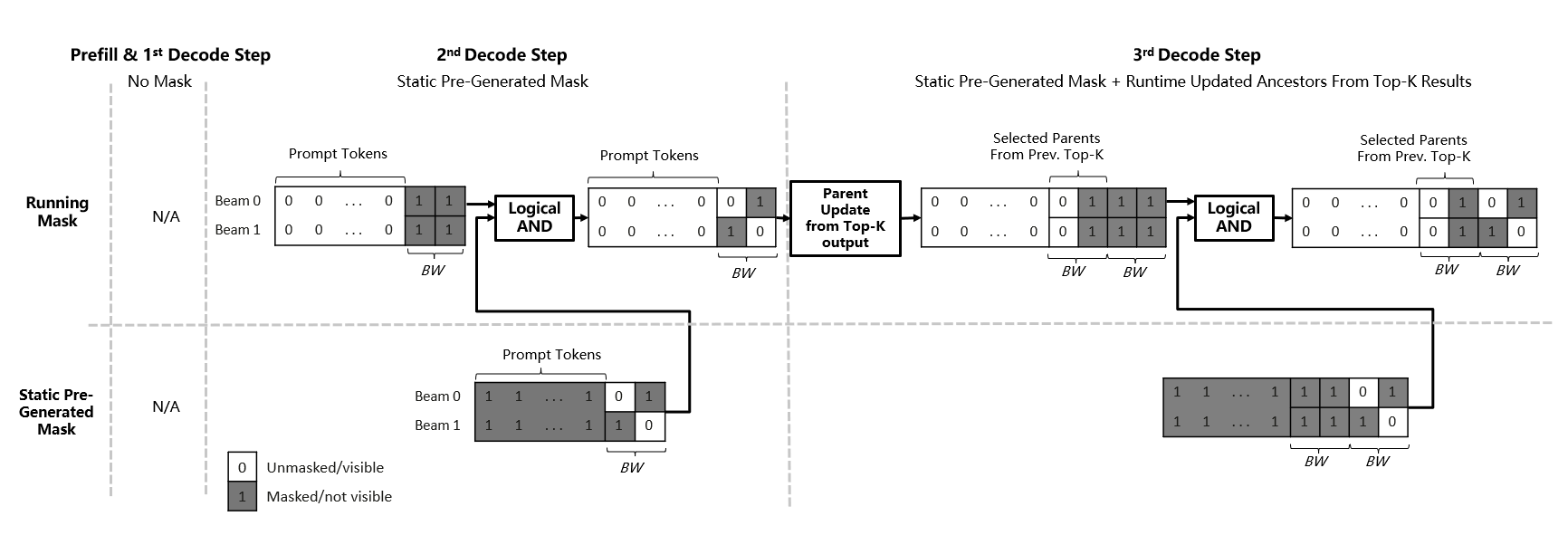}
\caption{Mask generation for the three-token generation loop in our Tree Attention implementation with a beam width of 2. Following the Ascend FlashAttention operator's masking convention, 0 denotes a visible element and 1 denotes a masked element. The parent update shown in this figure is based off the parent selection from Fig.~\ref{fig:treeatten}.} 
\label{fig:masking-example}
\end{figure*}

Constructing the attention mask during decoding introduces repeated tensor initialization, indexing, and comparison operations, motivating mask pre-generation before decoding. However, the complete mask cannot be fully pre-generated because the ancestor paths depend on runtime top-$k$ beam selection. We therefore decompose mask construction into a \verb|static mask| and a \verb|running mask|. The static mask represents the predefined visibility pattern of each decoding step and is pre-generated for every supported prompt length and beam width, while the running mask carries previously generated masks across decoding steps and updates each beam’s ancestors according to the parent indices selected at runtime. The two masks are combined through a lightweight \verb|logical AND| operation. This efficiently preserves previously visible positions while exposing the newly valid positions of the current decoding step, avoiding full mask reconstruction at runtime.

Figure~\ref{fig:masking-example} illustrates the mask-update process using the beam search example in Fig.~\ref{fig:treeatten}. During prefill and the first decoding step, the prompt is processed using standard causal attention without an explicit beam-ancestry mask. At the second decoding step, the initialized running mask is combined with the corresponding static mask through an element-wise \verb|logical AND| operation, producing the attention mask that is subsequently used in FlashAttention. Following the example in Fig.~\ref{fig:treeatten}, at the end of the second decoding step, Tokens E and F are selected, both descending from parent A. The running mask is therefore updated to unmask Token A while masking Token B. This update is performed through a single gather operation on the mask using parent indices from top-$k$ selection, allowing each beam’s ancestor visibility to be updated in parallel. At the third decoding step, the updated running mask is again combined with the static mask through \verb|logical AND| to produce the final mask passed to the FlashAttention operator.

The memory overhead for mask pre-generation is low. For a prompt length of 1024 and beam width of 512, the static mask occupies 4 MB of HBM, while the running mask requires 1 MB per active request, which is negligible relative to model weights and KV cache.

\subsection{Static Cache Initialization}
To address our second bottleneck of runtime KV block management overhead, we statically allocate a memory region of size $P + BW \times (N -1)$ for the KV states before graph capture, as shown in Fig.~\ref{fig:treeatten}, where $P$ is the prompt length, $BW$ is the beam width, and $N - 1$ is the number of decoding steps minus one, as the last token's KV values do not need to be cached. Since these dimensions are fixed for GR inference, the cache layout and per-step write regions are predetermined, eliminating runtime block allocation, resizing, and block-table management.

Our use of an attention mask removes the need to reorganize KV states after top-$k$ beam selection. All generated KV states remain at their original cache positions, including those from subsequently pruned branches, while the attention mask is updated to prevent future tokens from accessing invalid paths. This replaces costly KV state rearrangement after each decoding step for each transformer layer with a fast mask update that is common to all layers.

\subsection{Full Three-Token Loop Unrolling}

\begin{figure}[htbp]
    \centering

    \begin{subfigure}[t]{\linewidth}
        \centering
        \includegraphics[width=\linewidth]
            {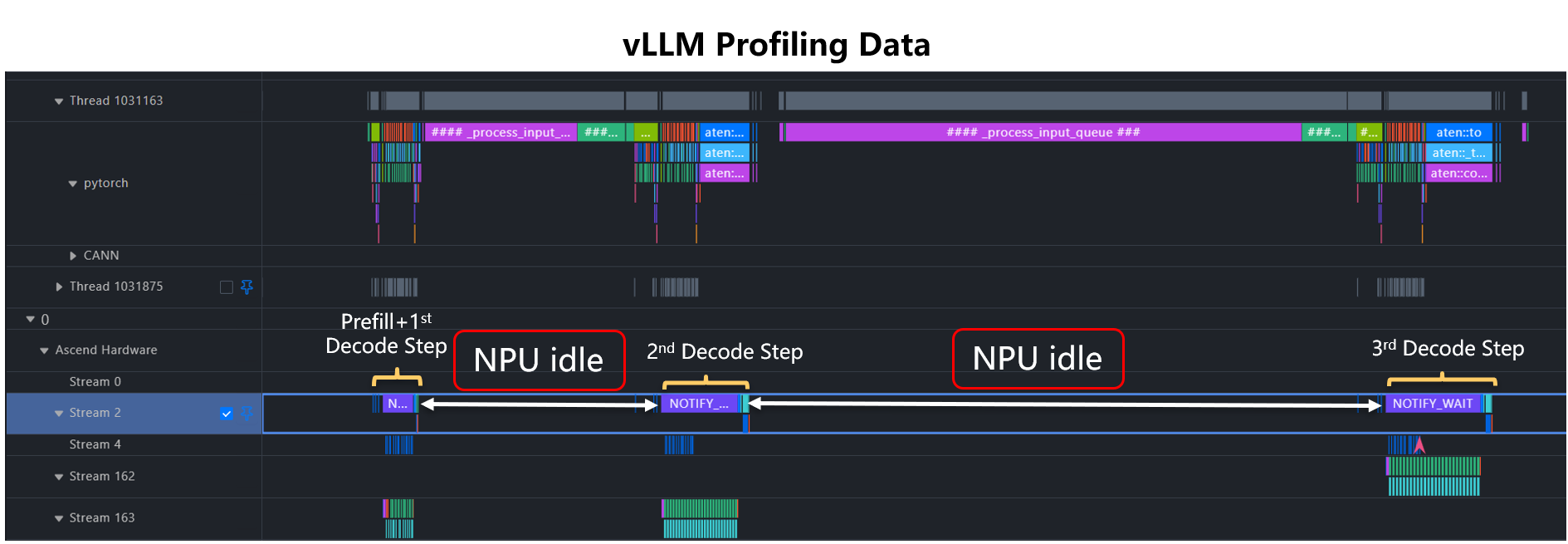}
        \caption{Profiling trace of vLLM showing NPU idle gaps between generation steps.}
        \label{fig:vllm_execution_timeline}
    \end{subfigure}

    \vspace{0.5em}

    \begin{subfigure}[t]{\linewidth}
        \centering
        \includegraphics[width=\linewidth]
            {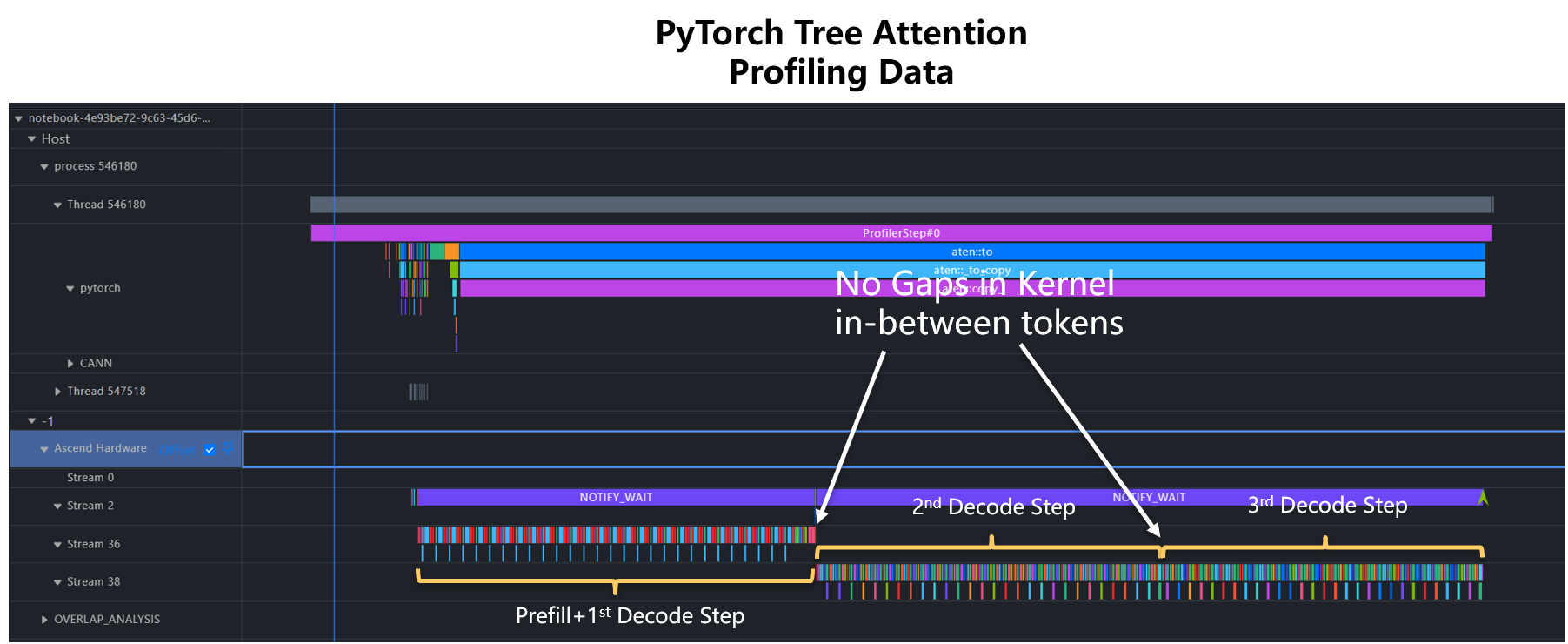}
        \caption{Profiling trace of our solution showing that loop unrolling eliminates NPU idle gaps between generation steps.}
        \label{fig:treeattention_execution_timeline}
    \end{subfigure}

    \caption{NPU profiled execution timelines for three-token generation.}
    \label{fig:execution_timeline_comparison}
\end{figure}

To address the third bottleneck of host-side beam processing, we exploit the fixed generation length in GR and unroll the entire decoding loop into a single NPU execution graph. Intermediate candidate scores, beam indices, attention masks, and KV cache states remain on the NPU, allowing beam selection and state updates to proceed without returning control to the host between decoding steps. This eliminates repeated engine invocations and host--device synchronization while exposing the complete fixed computation for further graph-level optimization.

Figure~\ref{fig:execution_timeline_comparison} compares the execution timelines of vLLM and our implementation. In vLLM, each decoding step requires a separate engine invocation. After each step, control returns to the host for beam selection, request construction, scheduling, and KV cache management before the next step can begin. These dependencies introduce substantial NPU idle gaps between decoding steps, particularly at large beam widths. In contrast, our loop-unrolled execution keeps the complete three-step generation on the NPU, effectively eliminating inter-step idle time.

\section{Further Optimizations}

\subsection{TorchAir Graph Compilation and Automatic Operator Fusion}
The fixed execution path of GR inference provides an opportunity for graph-level compilation and dispatch, replacing individual operator dispatches and reducing operator-launch overhead. We compile the entire decoding computation graph for Ascend NPUs with TorchAir~\cite{torchair}, so that each request is executed as a single device-side graph. Dedicated graphs are compiled beforehand for predefined prompt-length buckets, and the closest matching graph is selected at runtime. Inputs are padded to the corresponding bucket size when necessary, enabling static tensor shapes and reusable compiled graphs at the cost of a small amount of redundant computation. This static execution structure further allows TorchAir to optimize invariant tensor shapes and memory addresses, and to specialize operator tiling and autotuning for each predefined tensor shape.

TorchAir provides various compiler optimization and kernel fusion features. We enable tensor-view optimization, graph-pattern optimization, and Unified Buffer fusion for our implementation. Tensor-view optimization simplifies eligible reshape and dimension-permutation sequences, avoiding unnecessary tensor copies. For example, this is used in to avoid unnecessary tensor materialization during Q/K/V preparation and mask processing. Graph-pattern optimization fuses compatible operator sequences into a combined operator without writing to HBM, such as fusing residual addition followed by RMS normalization. Unified Buffer fusion combines compatible operations into compiler-generated kernels, allowing intermediate results to remain in the NPU’s on-chip Unified Buffer and reducing kernel-launch overhead and reads/writes to HBM. The resulting implementation combines Ascend-provided fused kernels that are explicitly selected in our code with kernels automatically fused by the compiler from compatible operation sequences, without requiring custom kernel implementations.

\subsection{SID-restricted Decode}

In GR, only a small subset of tokens from the model's vocabulary are valid item related Semantic-ID tokens, so we replace the final output projection over the full vocabulary with an output projection over only the valid subset of SID tokens. The projection's corresponding matrix multiplication operation, beam scoring, log-softmax, and top-$k$ selection therefore process 8,192 candidates per beam instead of 176,384 for OpenOneRec, reducing the candidate dimension by approximately $21.5\times$, and therefore significantly lowering latency for these operations.

\section{Evaluation}

We evaluate our implementation on OpenOneRec-1.7B and OpenOneRec-8B~\cite{openonerec} using a single Ascend 910B4 NPU with \texttt{torch\_npu} 2.8.0 and TorchAir 7.3.0. Model weights, activations, and KV cache are stored in the model default \verb|BFloat16| precision. We compare against two inference serving implementations. The first is vanilla vLLM, using vLLM Ascend v0.13.0 with prefix caching enabled. vLLM Ascend has a fixed KV block size of 128 tokens. The second is a SOTA custom vLLM implementation for GR that integrates the xAttention operator introduced in Sec.~2C to reduce redundant KV accesses. This solution also moves beam selection to the NPU, avoiding host--device synchronization between decoding steps. We evaluate these solutions using inference latency, where compute graph compilation latencies and warm-up overheads are excluded. There is no generation accuracy loss observed across all three implementations.

\subsection{Latency Comparisons}

\begin{table*}[ht]
\centering
\caption{Latency results on an Ascend 910B4 NPU. All values are reported in milliseconds. E2E latency is the sum of TTFT and decode time. NPU time measures token generation time excluding NPU idle time.}
\label{tab:latency_results}

\begin{subtable}{\textwidth}
\centering
\caption{OpenOneRec-1.7B, beam width 128, input length 1,024.}
\label{tab:latency_17b_bw128}

\makebox[\linewidth][c]{%
\begin{tabular}{@{}lrrrrr@{}}
\toprule
Method & E2E & TTFT & Decode Time &
NPU TTFT & NPU TPOT \\
\midrule
vLLM & 244.7 & 36.5 & 208.2 & 32.5 & 28.8 \\
Custom vLLM  & 59.9 & 33.5 & \textbf{26.4} & 30.4 & \textbf{13.2} \\
\textbf{Our Solution} & \textbf{59.1} & \textbf{30.7} & 28.4 & \textbf{28.5} & 14.2 \\
\bottomrule
\end{tabular}%
}
\end{subtable}

\vspace{0.75em}

\begin{subtable}{\textwidth}
\centering
\caption{OpenOneRec-1.7B, beam width 512, input length 1,024.}
\label{tab:latency_17b_bw512}

\makebox[\linewidth][c]{%
\begin{tabular}{@{}lrrrrr@{}}
\toprule
Method & E2E & TTFT & Decode Time &
NPU TTFT & NPU TPOT \\
\midrule
vLLM & 1989 & 37.4 & 1952 & 32.8 & 89.1 \\
Custom vLLM & 80.1 & 32.3 & \textbf{47.8} & 29.3 & \textbf{23.9} \\
\textbf{Our Solution} & \textbf{79.0} & \textbf{30.9} & 48.1 & \textbf{28.5} & 24.0 \\
\bottomrule
\end{tabular}%
}
\end{subtable}

\vspace{0.75em}

\begin{subtable}{\textwidth}
\centering
\caption{OpenOneRec-1.7B, beam width 512, input length 10,880.}
\label{tab:latency_17b_bw512_long}

\makebox[\linewidth][c]{%
\begin{tabular}{@{}lrrrrr@{}}
\toprule
Method & E2E & TTFT & Decode Time &
NPU TTFT & NPU TPOT \\
\midrule
vLLM & 5090 & 380.1 & 4710 & 374.4 & 879.8 \\
Custom vLLM & \textbf{454.3} & \textbf{370.5} & \textbf{83.8} & \textbf{364.8} & \textbf{41.8} \\
\textbf{Our Solution} & 467.7 & 379.3 & 88.4 & 369.8 & 44.0 \\
\bottomrule
\end{tabular}%
}
\end{subtable}

\vspace{0.75em}

\begin{subtable}{\textwidth}
\centering
\caption{OpenOneRec-8B, beam width 128, input length 1,024.}
\label{tab:latency_8b_bw128}

\makebox[\linewidth][c]{%
\begin{tabular}{@{}lrrrrr@{}}
\toprule
Method & E2E & TTFT & Decode Time &
NPU TTFT & NPU TPOT \\
\midrule
vLLM & 384.9 & 116.0 & 268.9 & 110.7 & 58.9 \\
Custom vLLM & \textbf{175.0} & 102.5 & \textbf{72.5} & 97.9 & \textbf{36.2} \\
\textbf{Our Solution} & 175.4 & \textbf{98.7} & 76.7 & \textbf{96.3} & 38.2 \\
\bottomrule
\end{tabular}%
}
\end{subtable}

\vspace{0.75em}

\begin{subtable}{\textwidth}
\centering
\caption{OpenOneRec-8B, beam width 512, input length 1,024.}
\label{tab:latency_8b_bw512}

\makebox[\linewidth][c]{%
\begin{tabular}{@{}lrrrrr@{}}
\toprule
Method & E2E & TTFT & Decode Time &
NPU TTFT & NPU TPOT \\
\midrule
vLLM & 2144 & 115.0 & 2029 & 110.8 & 158.9 \\
Custom vLLM & \textbf{239.6} & 100.3 & \textbf{139.3} & \textbf{97.2} & \textbf{69.5} \\
\textbf{Our Solution} & 246.4 & \textbf{99.3} & 146.3 & 97.3 & 73.0 \\
\bottomrule
\end{tabular}%
}
\end{subtable}

\end{table*}

Table~\ref{tab:latency_results} compares the inference latency of the three implementations under various beam widths and input lengths. End-to-end (E2E) latency measures the overall prefill and decode latency, including CPU processing, host--device synchronization, and NPU execution. It consists of Time-to-First-Token (TTFT), the time required to generate the first token, and Decode Time, the latency of generating the remaining two tokens. NPU TTFT and NPU Time-per-Output-Token (TPOT) measure only NPU execution time, where NPU TTFT denotes the NPU latency of the first token and NPU TPOT denotes the average NPU latency per token for the remaining two tokens.

Both GR-specific implementations achieve significant reductions in E2E latency relative to vanilla vLLM. For example, for OpenOneRec-1.7B with a beam width of 128 and an input length of 1,024 tokens, Table~\ref{tab:latency_results}~(a) shows that the custom vLLM and our implementation achieve E2E latencies of 59.9~ms and 59.1~ms, respectively, compared with 244.7~ms for vanilla vLLM, a speedup of  approximately 4$\times$. These gains come primarily from reductions in decode time. Vanilla vLLM performs beam path management on the host, incurring host-side processing and host-device synchronization overhead between decoding steps. This is reflected in the large difference between the NPU execution time for the last two tokens and the total decode latency. In Table~\ref{tab:latency_results}~(a), this corresponds to ($28.8 \times 2 = 57.6$~ms) for NPU time for the last 2 tokens, and 208.2~ms for its decode time. By unrolling the generation loop into a single device-side graph dispatch using TorchAir and keeping beam updates on the NPU, our approach eliminates expensive data movement, reducing this overhead to a negligible level. For the same model, increasing the beam width to 512 raises the speedup of our approach over vanilla vLLM to over 25$\times$, as shown in Table~\ref{tab:latency_results}~(b).

Furthermore, our approach achieves E2E latency comparable to the custom vLLM solution, with a difference of less than 3\%. This demonstrates that, with system-level optimizations, we can achieve performance similar to a solution that uses custom operators.

\subsection{Component-Level Breakdowns}

\begin{table*}[t]
\centering
\caption{Operator latencies for OpenOneRec-1.7B inference with an input length of 1,024 tokens and a beam width of 512. Each entry reports the sum of NPU kernel latencies grouped by component.}
\label{tab:operator_flow_latency}
\small
\setlength{\tabcolsep}{4pt}
\begin{tabular}{@{}lrrr@{}}
\toprule
Component & vLLM & Custom vLLM & Our Solution \\
& (ms) & (ms) & (ms) \\
\midrule
Prefill \& 1st decoding step
                                  & 32.267 & 28.335 & 28.425 \\
\midrule
KV cache/mask initialization     & --\textsuperscript{*} & 0.684 & 0.072 \\
\midrule
\multicolumn{4}{@{}l}{\textit{2nd decoding step}} \\
\quad Token embedding             & 0.043 & 0.050 & 0.037 \\
\quad QKV projection              & 2.871 & 2.905 & 4.953 \\
\quad RoPE                        & 0.516 & 0.560 & 0.720 \\
\quad Tensor preprocessing        & 0.517 & 0.543 & 2.271 \\
\quad Attention
                                  & 45.647 & 3.724 & 4.449 \\
\quad Tensor postprocessing       & 1.361 & 1.352 & 1.152 \\
\quad Add \& norm                 & 0.659 & 0.670 & 0.798 \\
\quad FFN                         & 8.191 & 9.321 & 9.263 \\
\quad Add \& norm                 & 0.429 & 0.430 & 0.469 \\
\quad Beam scoring (Vocab proj., softmax \& top-$k$)
                                  & 12.243\textsuperscript{+} & 2.318 & 0.500 \\
\addlinespace
\midrule
Beam-path update \& reconstruction    & --\textsuperscript{*} & 2.686 & 0.078 \\
\midrule
\multicolumn{4}{@{}l}{\textit{3rd decoding step}} \\
\quad Token embedding             & 0.050 & 0.048 & 0.037 \\
\quad QKV projection              & 4.141 & 2.915 & 4.902 \\
\quad RoPE                        & 0.690 & 0.557 & 0.620 \\
\quad Tensor preprocessing        & 0.904 & 0.543 & 0.869 \\
\quad Attention
                                  & 68.837 & 3.718 & 5.342 \\
\quad Tensor postprocessing       & 1.955 & 1.350 & 0.897 \\
\quad Add \& norm                 & 0.994 & 0.671 & 0.802 \\
\quad FFN                         & 13.557 & 9.315 & 8.639 \\
\quad Add \& norm                 & 0.835 & 0.426 & 0.470 \\
\quad Beam scoring (Vocab proj., softmax \& top-$k$)
                                  & 12.264\textsuperscript{+} & 2.323 & 0.484 \\
\addlinespace
\midrule
Beam-path update \& reconstruction    & --\textsuperscript{*} & 0.971 & 0.060 \\
\midrule
\textbf{Summed Operator NPU time}
& \textbf{208.971} & \textbf{76.413} & \textbf{76.307} \\
\bottomrule
\end{tabular}
\par\smallskip
\begin{minipage}{\linewidth}
\footnotesize
\textsuperscript{*}vLLM performs KV cache initialization and beam-path update \& reconstruction on the host.\\
\textsuperscript{+}vLLM is adapted to execute \verb|torch.topk()| on the NPU rather than on the host in our evaluation.
\end{minipage}
\end{table*}

Table~\ref{tab:operator_flow_latency} shows a breakdown of the NPU execution time for OpenOneRec-1.7B with a beam width of 512 and an input length of 1,024 tokens. Both GR-specific implementations substantially reduce the attention kernel and beam scoring latencies relative to vanilla vLLM. Across the second and third decoding steps, attention kernel time decreases from 114.484~ms in vanilla vLLM to 7.442~ms in custom vLLM and 9.790~ms in our approach. This is achieved by avoiding redundant attention computations over the shared prompt across beams in beam search. Beam scoring latency also decreases from 24.507~ms to 4.640~ms in custom vLLM and 0.984~ms in our approach. Vanilla vLLM computes token probabilities over the full vocabulary, whereas custom vLLM restricts top-$k$ selection to valid SID tokens. Our implementation goes one step further and applies this restriction earlier, at the vocabulary projection stage, avoiding the computation of logits and probabilities for invalid tokens. 

Our approach has a higher attention kernel latency than the custom vLLM solution, but comparing attention kernel times alone does not reflect the additional cost of beam KV management that is unique to the chosen method of attention. Furthermore, each attention operator has associated preprocessing and postprocessing (i.e., transposes, slicing, and copies) to satisfy the layout requirements of the attention operator. Although our approach has higher attention kernel latency and surrounding processing overheads, we focus on minimizing KV cache operations, including KV cache initialization, beam-path update and reconstruction overheads. Custom vLLM uses xAttention, which requires an unshared KV cache to be initialized at runtime and updated between decoding steps to reflect the selected parent beams. After the final decoding step, the selected beam sequences must also be reconstructed from the shared and unshared KV caches. Initialization takes 0.684~ms, while beam-path updates and reconstruction take 2.686~ms and 0.971~ms after the second and third decoding steps, respectively. Together, these operations take 4.341~ms. In contrast, our implementation retains KV states in their existing positions and updates attention masks to reflect beam ancestry, requiring only 0.209~ms for mask construction, beam-path updates and reconstruction combined. This is achieved because updating attention masks requires substantially less data movement than reorganizing KV tensors. Accounting for attention kernels, surrounding processing, and KV/mask management, our solution takes a total of 15.189~ms and custom vLLM takes 15.571~ms.

These results highlight the importance of holistically optimizing the complete generation path, or the ``cob", before optimizing the attention ``kernel" in isolation. Our mask-based design offsets the slower attention kernel and tensor preparation overheads, resulting in similar attention-related latency. As future work, a custom Tree Attention kernel that supports the existing tensor layouts could potentially further improve performance.

\section{Conclusion}
This work presents a system-level optimization approach for GR inference on Ascend NPUs. By combining tree attention with efficient mask and KV cache management, loop-unrolled generation, graph compilation, and vocabulary-constrained decoding, our implementation reduces redundant KV accesses and host-side overhead. Without custom kernels, it achieves up to 25$\times$ lower end-to-end latency than vLLM and achieves performance comparable to a SOTA implementation using specialized operators. These results demonstrate that system-level optimization can deliver substantial gains while retaining flexibility across model configurations and decoding workflows.

\bibliographystyle{IEEEtran}
\bibliography{biblio}

@misc{sidrajput2023,
      title={Recommender Systems with Generative Retrieval}, 
      author={Shashank Rajput and Nikhil Mehta and Anima Singh and Raghunandan H. Keshavan and Trung Vu and Lukasz Heldt and Lichan Hong and Yi Tay and Vinh Q. Tran and Jonah Samost and Maciej Kula and Ed H. Chi and Maheswaran Sathiamoorthy},
      year={2023},
      eprint={2305.05065},
      archivePrefix={arXiv},
      primaryClass={cs.IR},
      url={https://arxiv.org/abs/2305.05065}, 
}

@misc{xgr,
      title={{xGR}: Efficient Generative Recommendation Serving at Scale}, 
      author={Qingxiao Sun and Tongxuan Liu and Shen Zhang and Siyu Wu and Peijun Yang and Haotian Liang and Menxin Li and Xiaolong Ma and Zhiwei Liang and Ziyi Ren and Minchao Zhang and Yifan Wang and Xinyu Liu and Ke Zhang and Hailong Yang and Depei Qian},
      year={2026},
      eprint={2512.11529},
      archivePrefix={arXiv},
      primaryClass={cs.LG},
      url={https://arxiv.org/abs/2512.11529}, 
}

@misc{openonerec,
      title={{OpenOneRec} Technical Report}, 
      author={Guorui Zhou and Honghui Bao and Jiaming Huang and Jiaxin Deng and Jinghao Zhang and Junda She and Kuo Cai and Lejian Ren and Lu Ren and Qiang Luo and Qianqian Wang and Qigen Hu and Rongzhou Zhang and Ruiming Tang and Shiyao Wang and Wuchao Li and Xiangyu Wu and Xinchen Luo and Xingmei Wang and Yifei Hu and Yunfan Wu and Zhanyu Liu and Zhiyang Zhang and Zixing Zhang and Bo Chen and Bin Wen and Chaoyi Ma and Chengru Song and Chenglong Chu and Defu Lian and Fan Yang and Feng Jiang and Hongtao Cheng and Huanjie Wang and Kun Gai and Pengfei Zheng and Qiang Wang and Rui Huang and Siyang Mao and Tingting Gao and Wei Yuan and Yan Wang and Yang Zhou and Yi Su and Zexuan Cheng and Zhixin Ling and Ziming Li},
      year={2026},
      eprint={2512.24762},
      archivePrefix={arXiv},
      primaryClass={cs.IR},
      url={https://arxiv.org/abs/2512.24762}, 
}

@misc{pagedattention,
      title={Efficient Memory Management for Large Language Model Serving with {PagedAttention}}, 
      author={Woosuk Kwon and Zhuohan Li and Siyuan Zhuang and Ying Sheng and Lianmin Zheng and Cody Hao Yu and Joseph E. Gonzalez and Hao Zhang and Ion Stoica},
      year={2023},
      eprint={2309.06180},
      archivePrefix={arXiv},
      primaryClass={cs.LG},
      url={https://arxiv.org/abs/2309.06180}, 
}

@misc{unisgr,
      title={{UniSGR}: Unified Framework for Semantic {ID} Generation and Ranking}, 
      author={Jiawei Sun and Jun Yang and Ziyue Guo and Dongyue Xu and Jianan Yan and Lifang Deng and Xiaoyi Zeng},
      year={2026},
      eprint={2607.04068},
      archivePrefix={arXiv},
      primaryClass={cs.IR},
      url={https://arxiv.org/abs/2607.04068}, 
}

@inproceedings{tiger,
author = {Rajput, Shashank and Mehta, Nikhil and Singh, Anima and Keshavan, Raghunandan and Vu, Trung and Heidt, Lukasz and Hong, Lichan and Tay, Yi and Tran, Vinh Q. and Samost, Jonah and Kula, Maciej and Chi, Ed H. and Sathiamoorthy, Maheswaran},
title = {Recommender systems with generative retrieval},
year = {2023},
publisher = {Curran Associates Inc.},
address = {Red Hook, NY, USA},
booktitle = {Proceedings of the 37th International Conference on Neural Information Processing Systems},
articleno = {452},
numpages = {17},
location = {New Orleans, LA, USA},
series = {NIPS '23}
}

@inproceedings{plum,
author = {He, Ruining and Heldt, Lukasz and Hong, Lichan and Keshavan, Raghunandan and Mao, Shifan and Mehta, Nikhil and Su, Zhengyang and Tsai, Alicia and Wang, Yueqi and Wang, Shao-Chuan and Yi, Xinyang and Baugher, Lexi and Cakici, Baykal and Chi, Ed and Goodrow, Cristos and Han, Ningren and Ma, He and Rosales, Romer and Soest, Abby Van and Tandon, Devansh and Wu, Su-Lin and Yang, Weilong and Zheng, Yilin},
title = {PLUM: Adapting Pre-trained Language Models for Industrial-scale Generative Recommendations},
year = {2026},
isbn = {9798400723070},
publisher = {Association for Computing Machinery},
address = {New York, NY, USA},
url = {https://doi.org/10.1145/3774904.3792802},
doi = {10.1145/3774904.3792802},
booktitle = {Proceedings of the ACM Web Conference 2026},
pages = {8093–8104},
numpages = {12},
location = {United Arab Emirates},
series = {WWW '26}
}

@misc{torchair,
  author       = {{Huawei}},
  title        = {{TorchAir}},
  year         = {2024},
  howpublished = {\url{https://gitee.com/ascend/torchair}}
}

\end{document}